\documentclass[11pt]{article}
\usepackage{coling2020}
\usepackage{times}
\usepackage{url}
\usepackage{latexsym}

\usepackage{multirow}
\usepackage{graphicx}
\usepackage{tabularx}
\usepackage{booktabs}
\usepackage{comment}

\usepackage{booktabs}
\usepackage{varwidth}

\title{Linguistic Profiling of a Neural Language Model}

\author{Alessio Miaschi\textsuperscript{$\star$ $\diamond$}, Dominique Brunato\textsuperscript{$\diamond$}, {\bf Felice Dell'Orletta\textsuperscript{$\diamond$}, Giulia Venturi\textsuperscript{$\diamond$}}
\\
  \textsuperscript{$\star$}Department of Computer Science, University of Pisa\\
  \textsuperscript{$\diamond$}Istituto di Linguistica Computazionale ``Antonio Zampolli'', Pisa \\ItaliaNLP Lab -- \textit{www.italianlp.it}\\
  \normalsize{\texttt{alessio.miaschi@phd.unipi.it}, \texttt{\{name.surname\}@ilc.cnr.it}}}

\date{}

\begin{document}
\maketitle
\begin{abstract}
In this paper we investigate the linguistic knowledge learned by a Neural Language Model (NLM) before and after a fine-tuning process and how this knowledge affects its predictions during several classification problems. We use a wide set of probing tasks, each of which corresponds to a distinct sentence-level feature extracted from different levels of linguistic annotation. We show that BERT is able to encode a wide range of linguistic characteristics, but it tends to lose this information when trained on specific downstream tasks. We also find that BERT's capacity to encode different kind of linguistic properties has a positive influence on its predictions: the more it stores readable linguistic information of a sentence, the higher will be its capacity of predicting the expected label assigned to that sentence.
\end{abstract}

\section{Introduction}
\label{intro}

\blfootnote{
    %
    % for review submission
    %
    %\hspace{-0.65cm}  % space normally used by the marker
    %Place licence statement here for the camera-ready version. See
    %Section~\ref{licence} of the instructions for preparing a
    %manuscript.
    %
    % % final paper: en-uk version 
    %
     \hspace{-0.65cm}  % space normally used by the marker
    This work is licensed under a Creative Commons 
     Attribution 4.0 International Licence.
     Licence details:
     \url{http://creativecommons.org/licenses/by/4.0/}.
    % 
    % % final paper: en-us version 
    %
    % \hspace{-0.65cm}  % space normally used by the marker
    % This work is licensed under a Creative Commons 
    % Attribution 4.0 International License.
    % License details:
    % \url{http://creativecommons.org/licenses/by/4.0/}.
}

Neural Language Models (NLMs) have become a central component in NLP systems over the last few years, showing outstanding performance and improving the state-of-the-art on many tasks \cite{peters2018deep,radford2018improving,devlin2019bert}. However, the introduction of such systems has come at the cost of interpretability %and explainability
and, consequently, at the cost of obtaining meaningful explanations when automated decisions take place.
% and, specifically, of understanding how linguistic predictors - that were common as features in earlier systems - are encoded in such models.

Recent work has begun to study these models in order to understand whether they encode %are able to learn 
linguistic phenomena even without being explicitly designed %forse meglio trained? 
to learn such properties \cite{marvin2018targeted,goldberg2019assessing,warstadt2019investigating}. Much of this work focused on the analysis and interpretation of attention mechanisms \cite{tang-etal-2018-analysis,jain-wallace-2019-attention,clark-etal-2019-bert} and on the definition of \textit{probing models} trained to predict simple linguistic properties from unsupervised representations. 

Probing models trained  on different contextual representations provided evidences that such models are able to capture a wide range of linguistic phenomena \cite{adi2016fine,perone2018evaluation,tenney2019you} and even to organize this information in a hierarchical manner \cite{belinkov2017evaluating,lin-etal-2019-open,jawahar2019does}. However, the way in which this knowledge affects the decisions they make when solving specific downstream tasks has been less studied.

In this paper, we extended prior work by studying the linguistic properties encoded by one of the most prominent NLM, BERT \cite{devlin2019bert}, and how these properties affect its predictions when solving a specific downstream task.
%,  using a suite of more than 80 probing tasks. 
% qui vedere se tenere 'several' perché abbiamo 10 task di classificazione o dire che è uno solo diviso in 10 "sotto-task".
We defined three research questions aimed at understanding: (i) what kind of linguistic properties are already encoded in a pre-trained version of BERT and where across its 12 layers; (ii) how the knowledge of these properties is modified after a fine-tuning process; (iii) whether this implicit knowledge %of these properties 
affects the ability of the model to solve a specific downstream task, i.e. Native Language Identification (NLI). %With this aim, we firstly perform a very large suite of probing tasks using %on
%DOMI: SPOSTIAMO QUESTA PARTE
%To answer the first two questions, we firstly perform a very large suite of probing tasks using %on
%the sentence representations extracted from the internal layers of BERT. Each of these tasks makes explicit a particular property of the sentence, from very shallow features (such as sentence length) to more complex aspects of morpho--syntactic and syntactic structure (such as the depth of the whole tree and of specific sub-trees in a sentence), thus making them as particularly suitable to assess the implicit linguistic knowledge encoded in a NLM at a deep level of granularity. %with respect to a wide spectrum of phenomena overing lexical, morpho-syntactic and syntactic structure. 
To tackle the first two questions, we adopted an approach inspired to the `linguistic profiling' methodology put forth by \newcite{vanHalteren:2004}, which assumes that wide counts of linguistic features automatically extracted from parsed corpora allow modeling a specific language variety and detecting how it changes with respect to other varieties, e.g. complex vs simple language, female vs male--authored texts, texts written in the same L2 language by authors with different L1 languages. 
Particularly relevant for our study, is that multi-level linguistic features have been shown to have a highly predictive role in tracking the evolution of learners' linguistic competence across time and developmental levels, both in first and second language acquisition scenarios \cite{lubetich2014,miaschi-etal-2020}.

%when leveraged by traditional learning models on a variety of text classification problems, all of which can be successfully tackled using formal, rather than content based aspects of a text: from the assessment of sentence complexity and text readability \cite{Collins-Thompson:2014}, to the identification of personal and sociodemographics traits of an author, such as his/her native language, gender, age etc. \cite{nguyen-etal-2016-survey} and to the prediction of the evolution of learners' linguistic competence across time \cite{miaschi-etal-2020}. %From this perspective, our approach can be considered as a particular implementation of the `linguistic profiling' methodology put forth by \newcite{vanHalteren:2004}, which assumes that wide counts of linguistic features automatically extracted from parsed corpora allow modeling a specific language variety and detecting in what way it changes with respect to other varieties, e.g. complex vs simple language, female vs male--authored texts, texts written in the same L2 language by authors with different L1 languages.
Given the strong informative power of these features to encode a variety of language phenomena across stages of acquisition, we assume that they can be also helpful to dig into the issues of interpretability of NLMs. In particular, we would like to investigate whether features successfully exploited to model the evolution of language competence can be similarly helpful in profiling how the implicit linguistic knowledge of a NLM changes across layers and before and after tuning on a specific downstream task. We chose the NLI task, i.e. the task of automatically classifying the L1 of a writer based on his/her language production in a learned language \cite{malmasi-etal-2017-report}. 
%Secondly, we investigate the type and degree of variations of linguistic information before and after fine-tuning the pre-trained model on 10 distinct  datasets used to solve Native Language Identification (NLI), i.e. the task of automatically classifying the L1 of a writer based on his/her language production in a learned language \cite{malmasi-etal-2017-report}. 
 As shown by \newcite{Cimino2018SentencesAD}, linguistic features play a very important role when NLI is tackled as a sentence--classification task rather than as a traditional document--classification task. 
%NLI can be addressed by exploiting only linguistic features extracted at sentence--level reaching comparable performance to those obtained by state--of--the--art models based on word embeddings \cite{malmasi-etal-2017-report}. 
This is the reason why we considered the sentence-level NLI classification as a task particularly suitable for probing the NLM linguistic knowledge.
%\textbf{perché è un task che per essere risolto è necessario che il modello codifichi un'ampia gamma di informazioni linguistiche e anche perché è un task basato sull'info estratta dalla sentence -come dimostrato da Cimino et al (2017) nonostante lo stato dell'arte è stato definito soltanto usando word embeddings (Shared Task cit)}
%vecchia versione: a fine-tuning process based on a Native Language Identification (NLI) downstream task. 
%vecchia versione: -base and 10 fine-tuned models obtained training BERT on as many Native Language Identification (NLI) tasks. 
Finally, we investigated whether and which linguistic information encoded by BERT is involved in discriminating the sentences correctly or incorrectly classified by the fine-tuned models. To this end, we tried to understand if the linguistic knowledge that the model has of a sentence affects the ability to solve a specific downstream task involving that sentence.

%vecchia versione: Adopting a suite of more than 80 probing tasks, we firstly perform
% We perform our experiments using a suite of more than 80 probing tasks, each of which corresponds to a specific/distinct sentence-level feature. We find that / We show that

%The remainder of the paper is organized as follows. We start by presenting some related works which are more closely related to our study (Sec. \ref{rel_work}) and in Section \ref{approach} we highlight the main novelties of our approach. We then describe in more details the data (Sec. \ref{data}), the probing tasks (Sec. \ref{prob_tasks}) and the models (Sec. \ref{model}) we used. Experiments and results are described in Section \ref{experiment_1}, \ref{experiment_2} and \ref{experiment_3}. To conclude, in Section \ref{conclusion} we summarize the main findings of the study.

\paragraph{Contributions}
In this paper: (i) we carried out an in-depth linguistic profiling of BERT's internal representations
%deep analysis of the implicit linguistic knowledge stored in BERT's internal representations and how it changes across layers using a wide suite of sentence-level probing tasks, corresponding to a wide spectrum of linguistic phenomena at different level of complexity;
% we verify the implicit linguistic knowledge stored in BERT's internal representations using a suite of more than 80 probing tasks corresponding to a wide range of linguistic phenomena at different level of complexity; 
(ii) we showed that contextualized representations tend to lose their precision in encoding a wide range of linguistic properties %general-purpose linguistic properties 
after a fine-tuning process; % RIVEDERE 'GENERAL-PURPOSE' COME TERMINE PER DESCRIVERE LE NOSTRE FEATURES
(iii) we showed that the linguistic knowledge stored in the contextualized representations of BERT positively affects its ability to solve NLI downstream tasks: the more BERT stores information about these features% in its embeddings/internal representations
, the higher will be its capacity of predicting the correct label. 

\section{Related Work}
\label{rel_work}

% citare "Correlating neural and symbolic representations of language"
In the last few years, several methods have been devised to obtain meaningful explanations regarding the linguistic information encoded in NLMs \cite{belinkov2019analysis}. They range from techniques to examine the activations of individual neurons \cite{karpathy2015visualizing,li2016visualizing,kadar2017representation} to more domain specific approaches, such as interpreting attention mechanisms \cite{raganato2018analysis,kovaleva-etal-2019-revealing,vig-belinkov-2019-analyzing}, studying correlations between representations \cite{saphra2019understanding} or designing specific \textit{probing tasks} that a model can solve only if it captures a precise linguistic phenomenon using the \textit{contextual} word/sentence embeddings of a pre-trained model as training features \cite{conneau2018you,zhang2018language,hewitt2019designing,miaschi-dellorletta-2020-contextual}. These latter studies demonstrated that NLMs are able to encode a variety of language properties in a hierarchical manner \cite{belinkov2017evaluating,blevins2018deep,tenney2019you} and even to support the extraction of dependency parse trees \cite{hewitt2019structural}. \newcite{jawahar2019does} investigated the representations learned at different layers of BERT, showing that lower layer representations are usually better for capturing surface features, while embeddings from higher layers
are better for syntactic and semantic properties. Using a suite of probing tasks, \newcite{tenney-etal-2019-bert} found that the linguistic knowledge encoded by BERT through its 12/24 layers follows the traditional NLP pipeline: POS tagging, parsing, NER, semantic roles and then coreference. \newcite{liu-etal-2019-linguistic}, instead, quantified differences in the transferability of individual layers between different models, showing that higher layers of RNNs (ELMo) are more task-specific (less general), while transformer layers (BERT) do not exhibit this increase in task-specificity.

\section{Our Approach}
\label{approach}

\begin{table*}[t]
\scriptsize
    \centering
\begin{tabular}{lp{7cm}p{4.5cm}l}
\hline
\textbf{Level of Annotation}  & \textbf{Linguistic Feature} & \textbf{Label} \\\hline
\multirow{3}*{Raw Text} & \multicolumn{2}{c}{{\bf Raw Text Properties (\textit{RawText})}}\\%\cline{2-3}
& Sentence Length & sent\_length \\ 
& Word Length & char\_per\_tok \\ \hline
\multirow{2}*{Vocabulary} &
 \multicolumn{2}{c}{{\bf Vocabulary Richness (\textit{Vocabulary})}}\\%\cline{2-3}
& Type/Token Ratio for words and lemmas & ttr\_form, ttr\_lemma\\\hline
\multirow{5}*{POS tagging} & \multicolumn{2}{c}{{\bf Morphosyntactic information (\textit{POS})}}\\
& Distibution of UD and language--specific POS & upos\_dist\_*, xpos\_dist\_* \\
& Lexical density & lexical\_density \\\cline{2-3}
& \multicolumn{2}{c}{{\bf Inflectional morphology (\textit{VerbInflection})}}\\
& Inflectional morphology of lexical verbs and auxiliaries & xpos\_VB-VBD-VBP-VBZ, aux\_* \\\hline
\multirow{16}*{Dependency Parsing} & \multicolumn{2}{c}{{\bf Verbal Predicate Structure (\textit{VerbPredicate})}}\\
& Distribution of verbal heads and verbal roots & verbal\_head\_dist, verbal\_root\_perc\\
& Verb arity and distribution of verbs by arity & avg\_verb\_edges, verbal\_arity\_* \\\cline{2-3}
& \multicolumn{2}{c}{{\bf Global and Local Parsed Tree Structures (\textit{TreeStructure})}}\\
& Depth of  the  whole  syntactic tree & parse\_depth \\
 & Average length of dependency links and of the longest link & avg\_links\_len, max\_links\_len\\
 & Average length  of prepositional chains and distribution by depth & avg\_prep\_chain\_len, prep\_dist\_* \\
 & Clause length & avg\_token\_per\_clause \\\cline{2-3}
 & \multicolumn{2}{c}{{\bf Order of elements (\textit{Order})}}\\
 & Relative order of subject and object & subj\_pre, obj\_post \\\cline{2-3}
  & \multicolumn{2}{c}{{\bf Syntactic Relations (\textit{SyntacticDep})}}\\
 & Distribution  of  dependency  relations & dep\_dist\_* \\\cline{2-3}
 & \multicolumn{2}{c}{{\bf Use of Subordination (\textit{Subord})}}\\
 & Distribution of subordinate and principal clauses & principal\_prop\_dist, subordinate\_prop\_dist \\
 & Average length of subordination chains and distribution by depth & avg\_subord\_chain\_len, subordinate\_dist\_1 \\
 & Relative order of subordinate clauses & subordinate\_post \\\hline
 \end{tabular}
 \caption{Linguistic Features used in the probing tasks.}\label{LingFeat}

\end{table*}

%\begin{figure}
%\begin{center}
%\includegraphics[scale=1.7]{prob_diagram.png}
%\caption{Probing tasks process for a single sentence of the UD/NLI dataset. \textit{N} and \textit{M} correspond to the number of layers of the model and the number of linguistic features respectively.}
%\label{fig:prob_tasks_scheme}
%\end{center}
%\end{figure}
To probe the linguistic knowledge encoded by BERT and understand how it affects its predictions in several classification problems, we relied on a suite of 68 probing tasks, each of which corresponds to a distinct feature capturing lexical, morpho--syntactic and syntactic properties of a sentence. Specifically, we defined three sets of experiments. The first consisted in probing the linguistic information learned by a pre-trained version of BERT (BERT-base, cased) using gold sentences annotated according to the Universal Dependencies (UD) framework \cite{nivre2016universal}. In particular, we defined a probing model that uses BERT contextual representations for each sentence of the dataset and predicts the actual value of a given linguistic feature across the internal layers. %(See Figure \ref{fig:prob_tasks_scheme}). 
The second set of experiments consisted in investigating variations in the encoded linguistic information between the pre-trained model and 10 different fine-tuned ones obtained training BERT on as many Native Language Identification (NLI) binary tasks. To do so, we performed again all probing tasks using %the contextual representations of each sentence in the UD dataset extracted with 
the 10 fine-tuned models.
For the last set of experiments, we investigated how the linguistic competence contained in the models affects the ability of BERT to solve the NLI downstream tasks.

%In what follows, we introduce the datasets we use for our experiments (Sec. \ref{data}), the set of linguistic features for the definition of the probing tasks (Sec. \ref{prob_tasks}) and the two models (Sec \ref{model}).

\subsection{Data}
\label{data}
We used two datasets: (i) the UD English treebank (version 2.4) for probing the linguistic information learned before and after a fine-tuning process; (ii) a dataset  used for the NLI task, which is exploited both for fine-tuning BERT on the downstream task and for reproducing the probing tasks in the third set of experiments. % taking into account the performances of the fine-tuned models.

\paragraph{UD dataset} %In order to make the results of our probing tasks comparable, 
It includes three UD English treebanks: UD\_English-ParTUT, a conversion of a multilingual parallel treebank consisting of a variety of text genres, including talks, legal texts and Wikipedia articles \cite{sanguinetti2015parttut}; the Universal Dependencies version annotation from the GUM corpus \cite{Zeldes2017}; the English Web Treebank (EWT), a gold standard universal dependencies corpus for English \cite{silveira2014gold}. Overall, the final dataset consists of 23,943 sentences.

\paragraph{NLI dataset} We used the 2017 NLI shared task dataset, i.e.\ the TOEFL11 corpus \cite{blanchard2013toefl11}. It contains test responses from 13,200 test takers (one essay and one spoken response transcription per test taker) and includes 11 native languages (L1s) with 1,200 test takers per L1.  %To conduct separate series of experiments with different pairs of languages, 
We selected only written essays and we created pairwise subsets of essays written by Italian L1 native speakers and essays for all the other languages. At the end of this process, we obtained 10 datasets of 2,400 documents (33,756 sentences in average): 1,200 for the Italian L1 speakers and 1,200 for each of the other L1s included in the TOEFL11 corpus.

% vedere se mettere tabellina con numeri (frasi e doc in italiano, frasi e doc altra lingua e totale)

% ricontrollare ref per EWT

\subsection{Probing Tasks and Linguistic Features}
\label{prob_tasks}

Our experiments are based on the probing tasks approach defined in \newcite{conneau2018you}, which aims to capture linguistic information from the representations learned by a NLM. In our study, each probing task consists in predicting the value of a specific linguistic feature automatically extracted from the parsed sentences in the NLI and UD datasets. The set of features is based on the ones described in \newcite{profilingud-brunato-2020} % and counts more than 150 characteristics
 which are acquired from raw, morpho-syntactic and syntactic levels of annotation and can be categorised in 9 groups corresponding to different linguistic phenomena. As shown in Table \ref{LingFeat}, these features model linguistic phenomena ranging from raw text ones%(e.g.\ the average length of words and sentence)
 , to morpho--syntactic information and inflectional properties of verbs, to more complex aspects of sentence structure modeling global and local properties of the whole parsed tree and of specific subtrees, such as the order of subjects and objects with respect to the verb, the distribution of UD syntactic relations, also including features referring to the use of subordination and to the structure of verbal predicates.

\subsection{Models}
\label{model}
% fare breve intro ai modelli?
\paragraph{NLM} We relied on the pre--trained English version of BERT (BERT-base cased, 12 layers, 768 hidden units) for both the extraction of contextual embeddings and the fine-tuning process for the NLI downstream task. To obtain the embeddings representations for our sentence-level tasks we used for each of its 12 layers the activation of the first input token (\textit{[CLS]}), which somehow summarizes the information from the actual tokens, as suggested in \newcite{jawahar2019does}.
\paragraph{Probing model} As mentioned above, each of our probing tasks consists in predicting the actual value of a given linguistic feature given the inner sentence representations learned by a NLM for each of its layers. Therefore, we used a linear Support Vector Regression (LinearSVR) as probing model.

\section{Profiling BERT}
\label{experiment_1}

\begin{table}[]
\centering
\scriptsize
\begin{tabular}{lllllllllll}
\hline
\textbf{Models}   & \textbf{RawText} & \textbf{Vocabulary} & \textbf{POS}  & \textbf{VerbInflection} & \textbf{VerbPredicate} & \textbf{TreeStructure} & \textbf{Order} & \textbf{SyntacticDep} & \textbf{Subord} & \textbf{All} \\
\hline
BERT     & 0.68    & 0.78       & 0.68 & 0.72           & 0.60          & 0.78          & 0.72  & 0.69         & 0.71          & 0.69        \\
Baseline & 0.52    & 0.23       & 0.27 & 0.35           & 0.48          & 0.70          & 0.51  & 0.34         & 0.48          & 0.38 \\
\hline
\end{tabular}
\caption{BERT $\rho$ scores (average between layers) for all the linguistic features (\textit{All}) and for the 9 groups corresponding to different linguistic phenomena. Baseline scores are also reported.}
\label{table:avg_layers}
\end{table}

\begin{figure}[t]
\begin{center}
\includegraphics[width=0.65\textwidth]{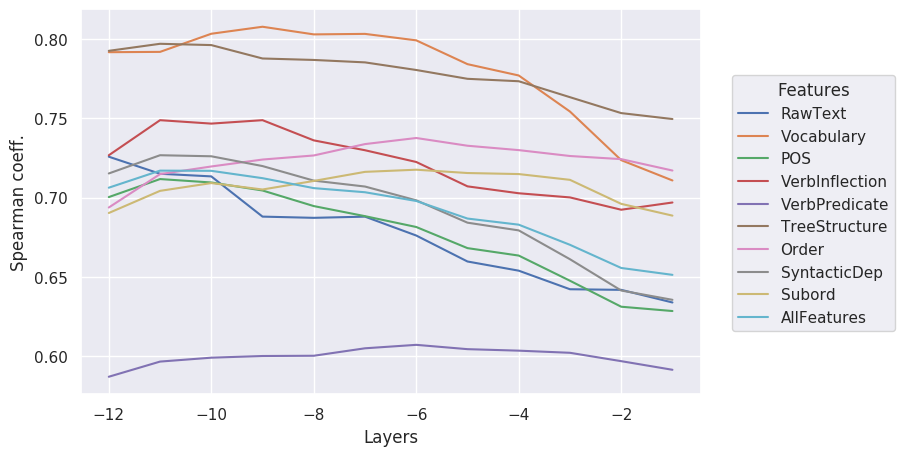}
\caption{BERT average layerwise $\rho$ scores.}% computed according to all the linguistic features (\textit{AllFeatures}) and according to the 9 groups corresponding to different linguistic phenomena.}
\label{fig:avg_layers}
\end{center}
\end{figure}

\begin{figure*}[t]
\begin{center}
\includegraphics[width=1\textwidth]{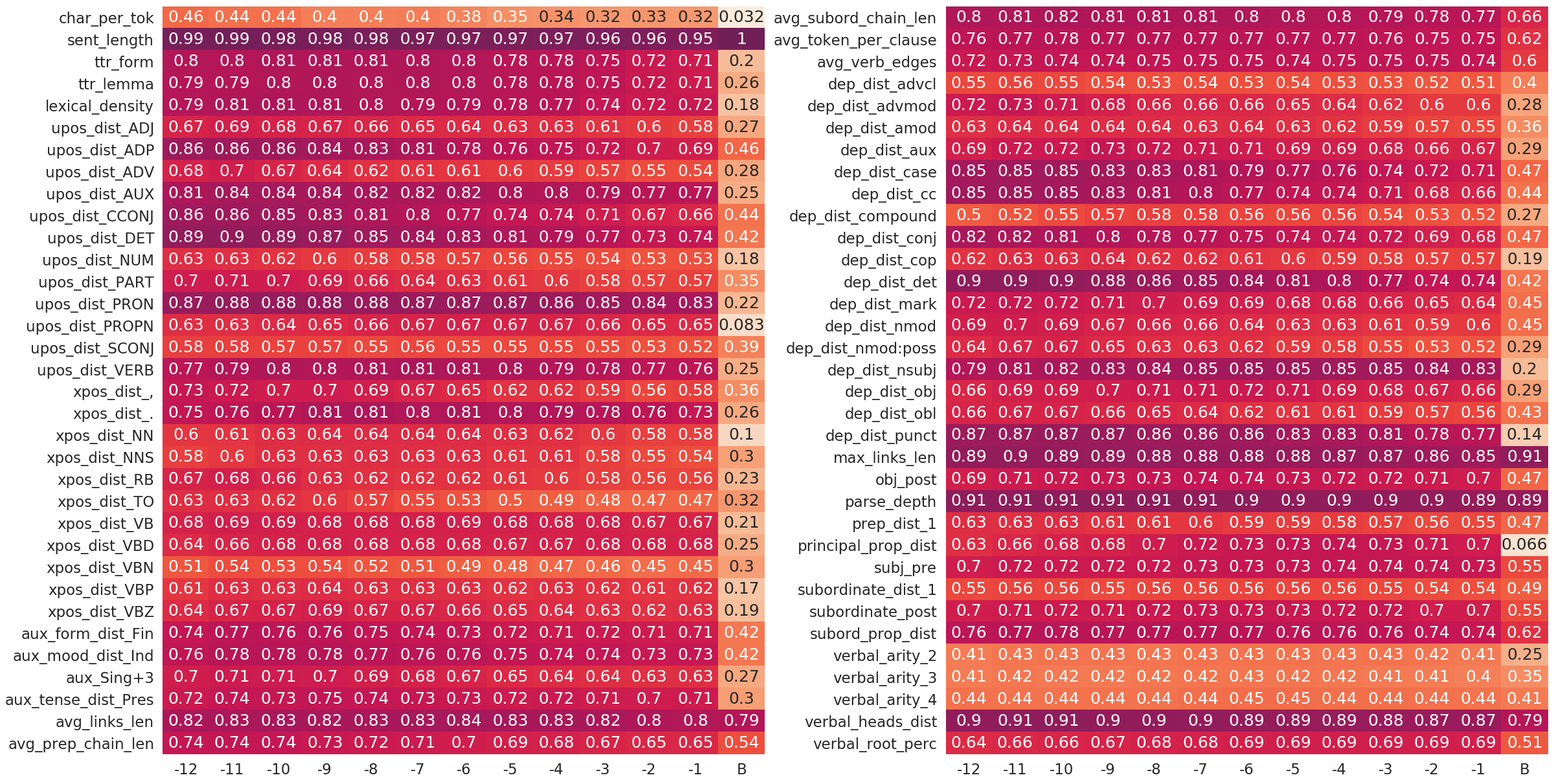}
\caption{Layerwise $\rho$ scores for the 68 linguistic features. Absolute baseline scores are reported in column \textit{B}.}
\label{fig:heatmap_probes}
\end{center}
\end{figure*}

Our first experiments investigated what kind of linguistic phenomena are encoded in a pre-trained version of BERT. To this end, for each of the 12 layers of the model (from input layer \textit{-12} to output layer \textit{-1}), we firstly represented each sentence in the UD dataset using the corresponding sentence embeddings according to the criterion defined in Sec.\ \ref{model}. We then performed for each sentence representation our set of 68 probing tasks using the LinearSVR model. %To probe BERT linguistic competence
Since most of our probing features are strongly correlated with sentence length, we compared the probing model results with the ones obtained with a
baseline computed by measuring the Spearman's rank correlation coefficient ($\rho$) between the length of the UD dataset sentences and the corresponding probing values. The evaluation is performed with a 5-fold cross validation and using
Spearman correlation ($\rho$) between predicted and gold labels 
%R\textsuperscript{2} and Mean Squared Error (MSE) 
as evaluation metric. 

As a first analysis, we probed BERT's linguistic competence with respect to the 9 groups of probing features. Table \ref{table:avg_layers} reports BERT (average between layers) and baseline scores for all the linguistic features and for the 9 groups corresponding to different linguistic phenomena. As a general remark, we can notice that the scores obtained by BERT's internal representations always outperform the ones obtained with the correlation baseline. %, thus showing its capacity of implicitly encodes a wide spectrum of linguistic phenomena.
%In line with the reported average scores, 
For both BERT and the baseline, the best results are obtained for groups including features highly sensitive to sentence length. For instance, this is the case of syntactic features capturing global aspects of sentence structure (\textit{Tree structure}). 
%: the longer the sentence e.g.\ the higher the type/token ratio or the longer the syntactic dependency links or the deeper the whole syntactic tree.
However, differently from the baseline, the abstract representations of BERT are also very good at predicting features related to other linguistic information such as morpho-syntactic (\textit{POS}, \textit{Verb inflection}) and syntactic one, e.g.\ the structure of verbal predicate and the order of nuclear sentence elements (\textit{Order}). 

We then focused on how BERT's linguistic competence changes across layers. These results are reported in Figure \ref{fig:avg_layers}, where we see that the average layerwise $\rho$ scores are lower in the last layers both for all distinct groups and for all features together. As suggested by \newcite{liu-etal-2019-linguistic}, this could be due to the fact that the representations that are better-suited for language modeling (output layer) are also those that exhibit worse probing task performance, indicating that Transformer layers trade off between encoding general and probed  features. However, there are differences between the considered groups: competences about raw texts features (\textit{RawText}) and the distribution of POS are lost in the very first layers (by layer -10), while the knowledge about the order of subject/object with respect to the verb, the use of subordination, as well as features related to verbal predicate structure is acquired in the middle layers.   
%Andando a studiare le competenze di BERT tra i layer, Figure \ref{fig:avg_layers} reports average layerwise $\rho$ scores

%- tutti perdono competenze negli ultimi strati
%- alcune feats perdono da subito: raw, pos
%- migliorano più avanti: order, subordination e verb predicate 

Interestingly, if we consider how the knowledge of each feature changes across layers (Figure  \ref{fig:heatmap_probes}), we observe that not all features belonging to the same group have  an homogeneous behaviour. This is for example the case of the two features included in the \textit{RawText} group: word length (\textit{char\_per\_tok}) achieves quite lower scores across all layers with respect to the \textit{sent\_length} feature.
Similarly, the knowledge about POS differs when we consider more granular distinctions. For instance, within the broad categories of verbs and nouns, worse predictions are obtained by sub--specific classes of verbs based on tense, person and mood features (see especially past participle, \textit{xpos\_dist\_VBN}), and by inflected nouns both singular and plural (\textit{\_NN}, \textit{\_NNS}). 
Within the broad set of features extracted from syntactic annotation, we also see that different scores are reported for features referring e.g. to types of dependency relations: %: that is to say, the ability to recognize the POS introducing them can be exploited as a cue for their identification. 
those linking a functional POS to its head (e.g.\ \textit{dep\_dist\_case, dep\_dist\_cc, dep\_dist\_conj, dep\_dist\_det}) are better predicted than others relations, such as \textit{dep\_dist\_amod, advcl}.
Besides, within the \textit{VerbPredicate} group, lower $\rho$ scores are obtained by features encoding sub-categorization information about verbal predicates, such as the distribution of verbs by arity (\textit{verbal\_arity\_2,3,4}), which also remains almost stable across layers. %, or among the features including in the \textit{Subordination} group the knowledge about the distribution of subordinate chains by length (\textit{subordinate\_dist\_1}). 

\begin{figure*}[t]
\begin{center}
\includegraphics[width=1\textwidth]{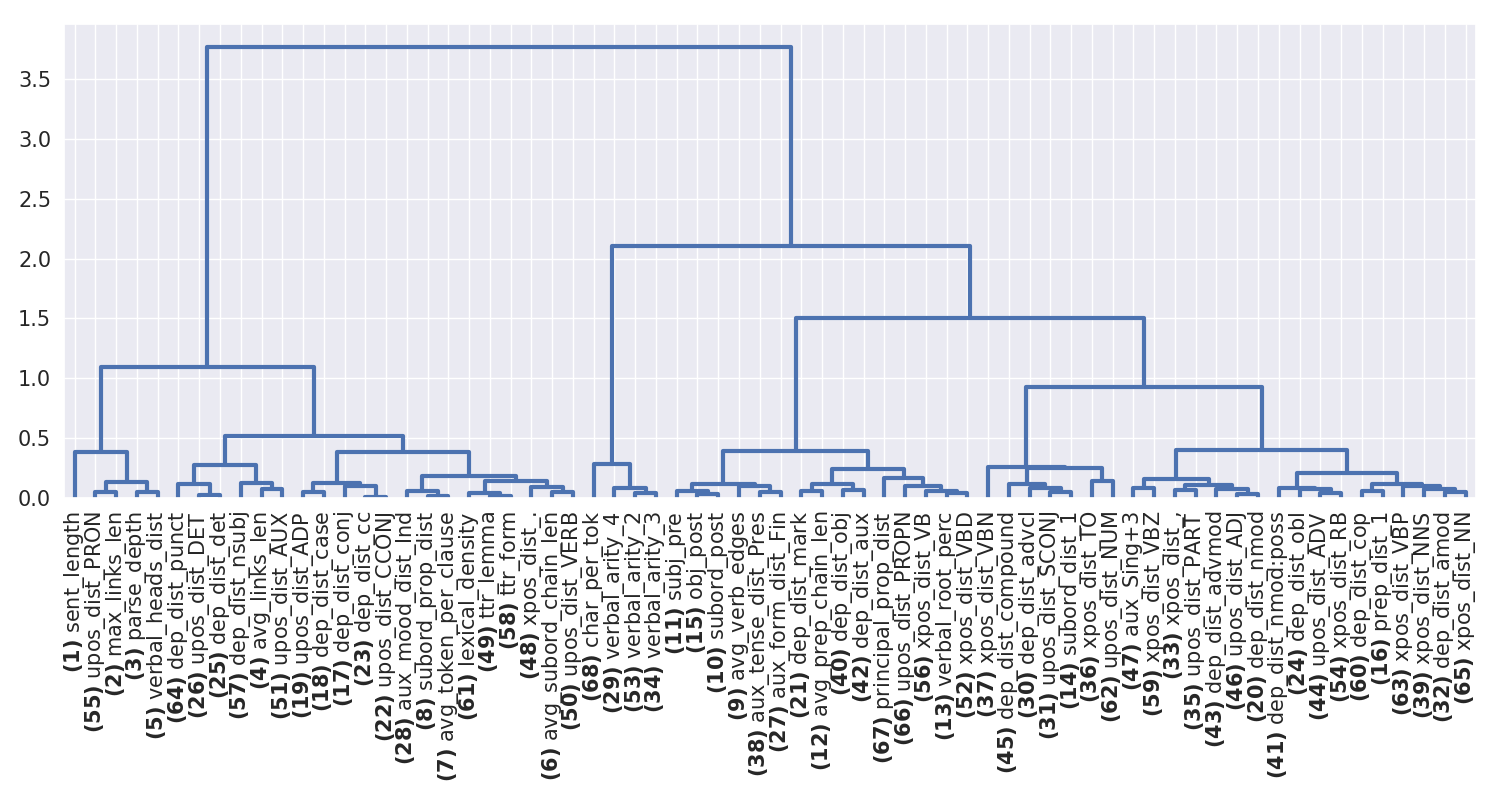}
\caption{Hierarchical clustering of the 68 probing tasks based on layerwise $\rho$ values. Bold numbers correspond to the ranking of each probing feature based on the correlation with sentence length.}
\label{fig:dendogram}
\end{center}
\end{figure*}

%In order to investigate how and where the NLM differently encodes features related to diverse typologies of language phenomena, we clustered our 68 linguistic characteristics according to layerwise probing results.
%In order to investigate which features are similarly encoded by the NLM, 
%

Since we observed these not homogeneous scores within the groups we defined a priori, we investigated how BERT hierarchically encodes across layers all the features.
%Avendo osservato i comportamenti non omogenei tra i gruppi, following \newcite{jawahar2019does} and \newcite{tenney-etal-2019-bert}, we then investigated if BERT hierarchically encodes across layers the wide spectrum of linguistic information keeping the traditional macro-classification in raw text, lexical, morpho-syntactic and syntactic features. 
To this end, we clustered the 68 linguistic characteristics according to layerwise probing results: specifically, we performed hierarchical clustering using Euclidean distance as distance metric and Ward variance minimization as clustering method. %In particular, we aimed at studying whether, in line with \cite{jawahar2019does} and \cite{tenney-etal-2019-bert}, BERT hierarchically encodes across layers the wide spectrum of linguistic information keeping the traditional macro-classification in raw text, lexical, morpho-syntactic and syntactic features.  
%Building on observations \cite{jawahar2019does,tenney-etal-2019-bert} that linguistic information is encoded in a hierarchical manner (lower layers for surface features and higher layers for more complex linguistic properties), we cluster our 68 linguistic characteristics according to layerwise probing results. 
Interestingly enough, Figure \ref{fig:dendogram} shows that the traditional division of features with respect to the linguistic annotation levels has not been maintained. On the contrary, BERT puts together  features from all linguistic groups into clusters of different size. In addition, these clusters gather features that are differently ranked according to the baseline scores (ranking positions are bolded in the figure). For example, the first cluster includes features with similar $\rho$ scores, and both highly and lower ranked by the baseline. %This result shows that BERT's representations encode with a similar accuracy information not only related to sentence length but also to different linguistic phenomena. More specifically, the first--level split highlights \textit{i)} a quite homogeneous cluster of features (encoded with similar $\rho$ scores) that 
All these features model aspects of global sentence structure, e.g.\ \textit{sent\_length}, functional POSs (e.g.\ \textit{upos\_dist\_DET}, \textit{\_ADP}, \textit{\_CCONJ}), parsed tree structures (e.g.\ \textit{parse\_depth, verbal\_heads\_dist, avg\_links\_len}), nuclear  elements of the sentence such as subjects (\textit{dep\_dist\_nsubj}), verbs (\textit{\_VERBS}), pronouns (\textit{\_PRON}). 
%as well as functional POSs (e.g.\ \textit{upos\_dist\_DET}, \textit{\_PRON}, \textit{\_ADP}) and parsing (e.g.\ \textit{parse\_depth, dep\_dist\_nsubj, verbal\_heads\_dist, avg\_links\_len}), 
%, and \textit{ii)} a cluster where all other features are mixed together. %including those referring to verbal predicate modification, encoded both in terms of tense and mood (\textit{xpos\_dist\_VBN}, \textit{verbs\_participle}) and of information extracted from the parse tree (\textit{verbal\_arity\_*}, \textit{subordinate\_dist\_1}, \textit{dep\_dist\_advcl,\_obl}), as well as word length (\textit{char\_per\_tok}). 
%Figure \ref{fig:clusters_results} reports the different R\textsuperscript{2} scores across clusters grouping linguistically-homogeneous phenomena.\\ 

%that linguistic information is encoded in a hierarchical manner (lower layers for surface features and higher layers for more complex linguistic properties), we cluster our 68 linguistic characteristics according to layerwise probing results.

%\textbf{Commentare il dendogramma raggruppando le features dall'alto verso il basso e osservandone l'andamento per strati nei vari plot.}

% The assumption is that if BERT solves probing tasks belonging to the same linguistic category (lexical, morpho-syntax, syntax) with similar results, e.g. higher scores in lower layers for surface features, then it is likely that this macro-categories' distinction will be maintained within the automatically extracted clusters. 

\section{The Impact of Fine--Tuning on Linguistic Knowledge}
\label{experiment_2}

\begin{table}[t]
\small
    \centering
     \begin{tabular}{ccccccccccc}
     \hline
    & \textbf{KOR} & \textbf{TEL} & \textbf{HIN} & \textbf{JPN} & \textbf{CHI} & \textbf{TUR} & \textbf{ARA} & \textbf{GER} & \textbf{FRE} & \textbf{SPA} \\
    \hline
    \textbf{Baseline} & 59.05   & 51.32   & 54.09   & 56.27   & 55.68   & 55.66   & 52.92   & 59.29   & 56.03   & 52.61   \\
    \textbf{BERT} & 85.74   & 85.18   & 84.75   & 84.19   & 82.78   & 79.29   & 76.38   & 72.78   & 72.50   & 70.03 \\
    \hline
    \end{tabular}
    \caption{NLI classification results in terms of accuracy. We used the Zero Rule algorithm as baseline. Note that, for each task, sentences of the 10 languages are paired with the Italian ones (e.g. KOR = KOR-ITA).}
    \label{tab:nli_accuracies}
\end{table}

%The second set of experiments aims at investigating whether and how the linguistic knowledge encoded by BERT changes after a fine-tuning process.
Once we have probed the linguistic knowledge encoded by BERT across its layers,  we investigated how it changes after a fine-tuning process.
%To do so, we firstly fine-tuned the same pre-trained model that was used for our previous experiments on our target NLI task. In particular,
To do so, we started with the same pre-trained version of the model used in the previous experiment and performed a fine-tuning process for each of the 10 subsets built from the original NLI corpus (Sec. \ref{data}). %We fine-tuned BERT using the default hyperparameters given by Google (num. epochs = 3, learning rate = 2e\textsuperscript{-5}). 
We decided to use 50\% of each NLI subset for  training  (40\% and 10\% for training and development set) and the remaining 50\% for testing the accuracy of the newly generated models. 

Table \ref{tab:nli_accuracies} reports the results for the 10 binary NLI tasks.
As we can notice, BERT achieves good results for all downstream tasks, meaning that is able to discriminate the L1 of a native speaker on a sentence-level regardless of the L1 pairs taken into account. The best performance is achieved by the model that was fine-tuned on the Korean and Italian pairwise subset, %native language identification task (KOR-ITA)
while the lowest scores are obtained with the model trained on the subset consisting of essays written by Spanish and Italian L1 speakers (SPA-ITA). Interestingly, these results seem to reflect typological distances among L1 pairs, with higher scores for languages that are more distant from Italian (Korean, Telugu or Hindi) and lower scores for L1s belonging to the same language family (FRE-ITA or SPA-ITA).

\begin{figure}[t!]
\begin{center}
\includegraphics[width=0.47\textwidth]{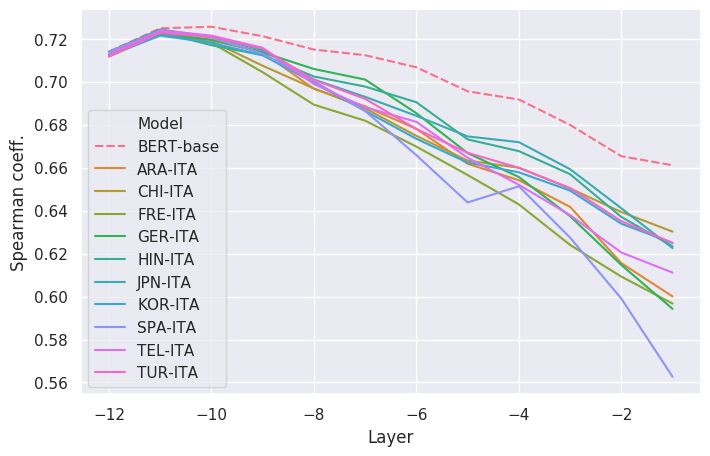}
\caption{Layerwise mean $\rho$ scores for the pre-trained and fine-tuned models.}
\label{fig:r2_lineplot}
\end{center}
\end{figure}

\begin{figure*}[t]
\begin{center}
\includegraphics[width=1.0\textwidth]{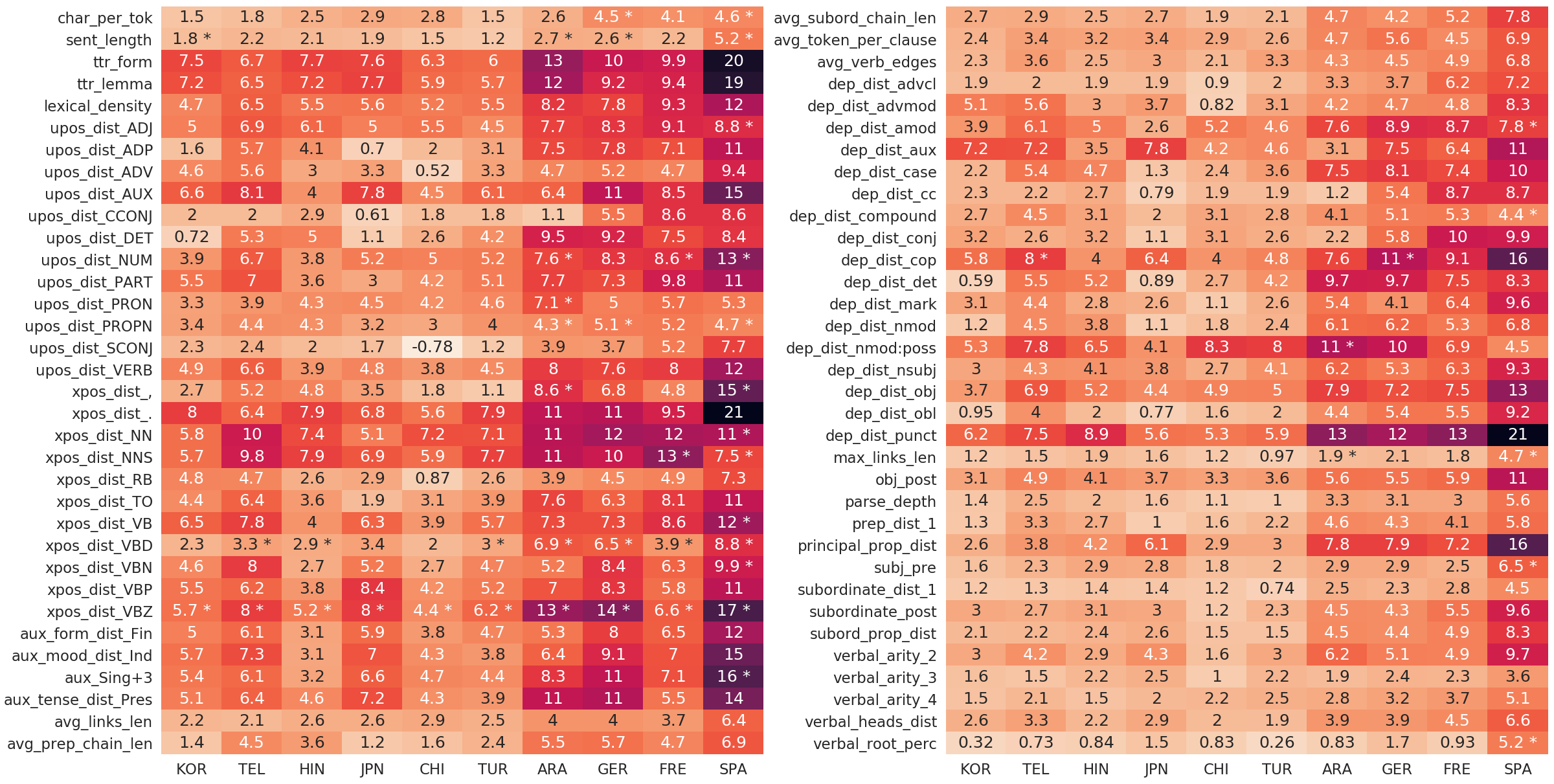}
\caption{Differences between BERT--base and fine--tuned models $\rho$ scores (multiplied by 100) computed using the output layer representations (\textit{-1}). Statistically significant variations (Wilcoxon Rank-sum test) are marked (\textit{*}).}
\label{fig:diff_heatmap}
\end{center}
\end{figure*}

After fine-tuning the model on NLI, we performed again the suite of probing tasks on the UD dataset using the 10 newly generated models and following the same approach discussed in Section \ref{experiment_1}. Figure \ref{fig:r2_lineplot} reports layerwise mean $\rho$ correlation values for all probing tasks obtained with BERT-base and the other fine-tuned models. %before and after fine-tuning BERT on the NLI tasks. 
It can be noticed that the representations learned by the NLM tend to lose their precision in encoding our set of linguistic features after the fine-tuning process. This is particularly noticeable at higher layers and it possibly suggests that the model is storing task--specific information at the expense of its ability to encode general knowledge about the language. Again, this is particularly evident for the models fine--tuned on the classification of language pairs belonging to the same family, SPA--ITA above all. To study which phenomena are mainly involved in this loss, we computed the differences between the probing tasks results obtained before and after the fine-tuning process. We focused in particular on the scores obtained on the output layer representations (layer \textit{-1}), since it is the most task-specific \cite{kovaleva-etal-2019-revealing}. For each subset, Figure \ref{fig:diff_heatmap} reports the difference between the score of each linguistic feature obtained with the pre--trained model and the fine--tuned one. %: the star marks features for which the difference is statistically significant using Wilcoxon Rank-sum test. 
Not surprisingly, the loss of linguistic knowledge reflects the typological trend observed for overall classification performance. In fact, when the task is to distinguish Italian vs German, French and Spanish L1, BERT loses much of its encoded knowledge for almost all the considered features. This is particularly evident for the morpho-syntactic features (i.e.\ distribution of \textit{upos\_dist} and \textit{xpos\_dist}) and for features related to lexical variety (i.e.\ \textit{ttr\_form}, \textit{ttr\_lemma}). It seems that for typologically similar languages BERT needs more task-specific knowledge mostly encoded at the level of morpho-syntactic information rather than the structural level. %It follows that there are few differences between the pre-training and fine-tuning scores obtained for the syntactic features. %({\bf Domi: non so se questa parte ci piace: la pensavo come un possibile aggancio alla terza sezione}). 
On the contrary, the drop is less pronounced and in most cases  not significant for models fine--tuned on the classification of more distant languages (e.g. models fine--tuned on KOR-ITA or TUR-ITA). %
%with respect to many syntactic features, e.g. \textit{verbal\_root, parse\_tree\_depth}).
In this case, the quite stable performance on the probing tasks may suggest that those features were still useful to perform the downstream task.
Interestingly, the class of features that decreases significantly in all models are those encoding the knowledge about the tense of verbs. This is particularly the case of the third-person singular verbs in the present tense (\textit{xpos\_dist\_VBZ}) and of verbs in the past tense (\textit{xpos\_dist\_VBD}). A possible explanation could be related to the prompts of essays, which are the same across the NLI dataset. Thus, the textual genre could have favored a quite homogeneous use of verbal morphology features by students of all L1s. This makes this class of features less useful for the identification of native languages.

\section{Are Linguistic Features useful for BERT's predictions?}
% Explaning BERT's predictions
\label{experiment_3}

\begin{figure*}[t]
\begin{center}
\includegraphics[width=0.96\textwidth]{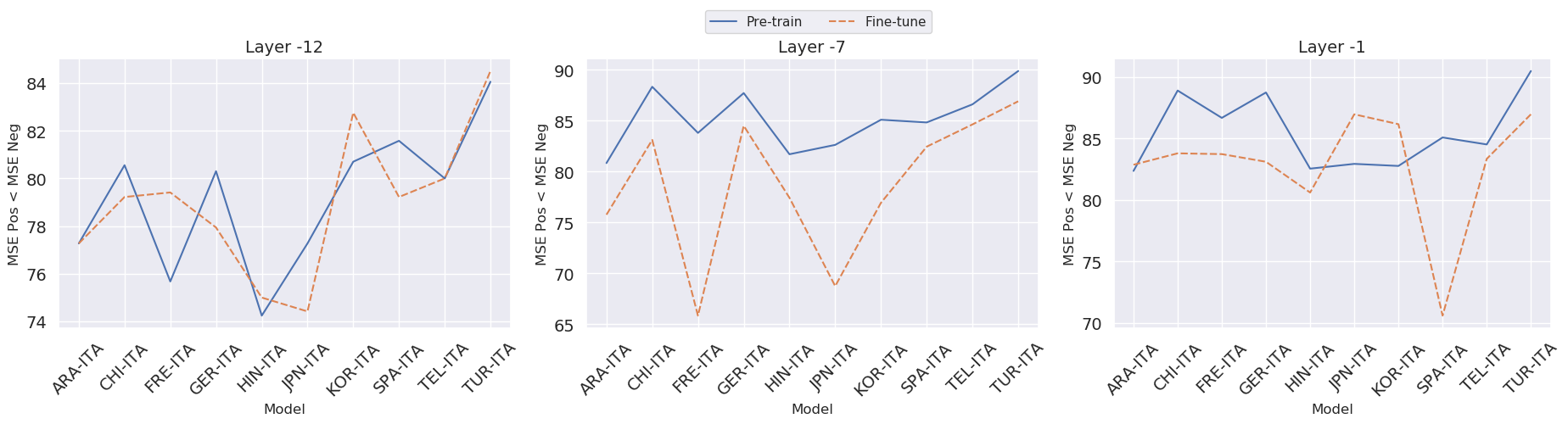}
\caption{\% of probing features for which the MSE of the sentences correctly classified by BERT-base (\textit{Pre-train}) and the fine-tuned models (\textit{Fine-tune}) is lower than that of the incorrectly ones. Results are reported for layers -12, -7 and -1.}
\label{fig:mse_plot}
\end{center}
\end{figure*}

%{\bf DOMI: per introdurre questa parte enfatizzerei di più il fatto che, nonostante abbiamo osservato nell'esperimento 2 che BERT in generale perde conoscenza, sono comunque molte le features che mantiene dopo il fine-tuning. Di conseguenza, ci chiediamo se queste gli possano servire anche quando svolge un task specifico}

As a last research question we investigated whether the implicit linguistic knowledge affects BERT's predictions when solving the NLI downstream task. %We focused in particular on the linguistic knowledge that varies significantly when BERT successfully solves the task. 
To answer this question we have split each NLI subset into two groups, i.e. sentences correctly  classified according to the L1 and those incorrectly classified. %pecifically, we use for each subset the remaining 50\% that we adopted for testing the accuracy of the corresponding model (Sec. \ref{experiment_2}). 
For the two groups of each NLI subset, we performed the probing tasks using the pre--trained BERT-base and the specific NLI fine-tuned model. %Specifically, we use for each subset the remaining 50\% that we adopted for testing the accuracy of the corresponding model (Sec. \ref{experiment_2}). Next, we split each NLI subset according to the sentences that were correctly or incorrectly classified and 
For each sentence of the two groups, we calculated the variation between the actual and predicted feature value obtaining two lists of absolute errors. We used the Wilcoxon Rank-sum test to verify whether the two lists were selected from samples with the same distribution.
%between the absolute errors of the two subgroups (sentences correctly and incorrectly classified) for each of our probing feature, we assessed how many vary in a statistically significant way. 
%Table \ref{tab:wilcoxon_feats} reports the results for the input (\textit{Layer -12}), middle (\textit{Layer -7}) and output (\textit{Layer -1}) layers and according to BERT-base (\textit{Pre}) and the 10 fine-tuned models (\textit{Fine}). 
As a general remark, we observed that much more than half of features vary in a statistically significant way between correctly and incorrectly classified sentences. This suggests that BERT's linguistic competence on the two groups of sentences is very different. %, with a lower number for JPN-ITA and FRE-ITA subsets. 
%Indipendentemente da layer e modelli, questa variazione è un po' superiore nel pre-trained e nei livelli centrali
%That is to say that many linguistic phenomena are involved in discriminating the sentences that were correctly and incorrectly classified by the models. More specifically, we observe some differences across layers and models. On the one hand, a higher number of features varies significantly in the input and output layers when sentences were classified by the fine-tuned models. On the other hand, the number of varying features in the middle layer is always higher using the BERT--based model, with the exception of FRE-ITA subset. 
%for the linguistic properties extracted using the internal representations learned by BERT-base. 
%Moreover, we notice that for JPN-ITA and FRE-ITA subsets the number of features that vary in a statistically significant way between correct and incorrect predictions is slightly lower with respect to the others.
To deepen the analysis of this difference, we calculated the accuracy achieved by BERT in terms of Mean Square Error (MSE) only for the set of features varying in a significant way.
%Figure \ref{fig:mse_plot} reports, only for the features that vary in a statistically significant way, the percentage of those achieving the best accuracy among 
%{\bf Per ogni feat che varia significativamente, la Figura 5 riporta the percentage of probing tasks che sono risolti meglio nelle frasi corrette rispetto a quelle sbagliate (where the accuracy is calculated as Mean Squared Error (MSE)). }
%The second analysis aims at investigating if the probing tasks achieve better results when BERT correctly solve the NLI task. 
%if BERT encodes more information about linguistic properties of sentences when it correctly predicts the L1 of a native speaker. %solving the downstream task, i.e. when predicting the correct label.
%To do so, we compute the mean squared error (MSE) for the probing features considered statistically significant in distinguishing between sentences that were correctly and incorrectly classified. 
%As we can see in 
Figure \ref{fig:mse_plot} reports %regardless of the layer and model taken into account, 
the percentage of features for which the MSE of the sentences correctly classified (\textit{MSE Pos}) is lower than that of the incorrectly ones (\textit{MSE Neg}). This percentage is significantly higher, thus showing that BERT's capacity to encode different kind of linguistic information could have an influence on its predictions: the more BERT stores readable linguistic information into the representations it creates, the higher will be its capacity of predicting the correct L1. Moreover, we noticed that this is true also (and especially) using the pre-trained model. In other words, this result suggests that the evaluation of the linguistic knowledge encoded in a pre--trained version of BERT on a specific input sequence could be an insightful indicator of its ability in analyzing that sentence with respect to a downstream task.
%This means that evaluating the linguistic knowledge of a pre--trained BERT on a given sentence could be an insightful indicator of its ability in analyzing that sentence with respect to a downstream task.

Interestingly, if we analyze the average length of correct and incorrect classified sentences, the correct ones are much more longer than the others for all tasks (from 3 tokens more for SPA-ITA to 9 for TEL-ITA). %This is quite expected for the NLI task, since more linguistic information makes this kind of tasks easier for a sentence classifier \textbf{(ALE: qui abbiamo qualcosa da citare?)}  
This is quite expected for the NLI task, since a higher number of linguistic events possibly occurring in longer sentences are needed to classify the L1 of a sentence \cite{dellorletta-etal-2014-assessing}. %On the contrary, longer sentences contain more complex syntactic structures \textbf{(ALE: anche qui qualche cit per avvalorare questa cosa?)}, which should increase the average complexity of the probing tasks. {\bf GIULIA: introdurrei questa frase dicendo: 
%Allo stesso tempo frasi più lunghe permettono una variazione maggiore rispetto alle probing features rendendo i probing task stessi più complessi
At the same time, longer sentences make more complex the probing tasks because the output space is larger for almost all them.
%At the same time, it is also well acknowledged that longer sentences may contain complex linguistic structures, such as longer dependencies and higher syntactic trees \cite{Gibson98linguisticcomplexity:}: 
%this should increase the average complexity of the probing tasks. %\textbf{[ Domi: sì, e forse anche qualcosa molto generico sulla complessità come perplessità...: ogni nuova parola espande lo spazio di ricerca, i potenziali attaccamenti ecc.?]}
%On the contrary, it is very surprising for the probing tasks, given that more tokens in a sentence increase the possible values that each linguistic features can have, potentially making the probing task more difficult. 
%This is an additional evidencne that BERT's linguistic knowledge is much higher not for simple sentences but for sentences that it will be able to better classify.  
This is an additional evidence that BERT's linguistic knowledge is not strictly related to sentence complexity, but rather to the model's ability to solve a specific downstream task. To confirm this hypothesis and verify whether such tendency does not only depend on sentence length, we trained another LinearSVR that takes as input the sentence length and predict our probing tasks according to correctly or incorrectly classified NLI sentences. %Results, reported in Table \ref{tab:sent_length}, showed that sentences incorrectly classified are more correlated with sentence length with respect to the correct ones, meaning that, in line with what we stated above, the capacity of BERT in encoding implicit linguistic linguistic knowledge and consequently in correctly classifying the NLI sentences is not necessarily related to sentence complexity. \textbf{Da espandere e/o chiarire meglio.} {\bf GIULIA: forse si puo' girare in questo modo: 
Table \ref{tab:sent_length} reports the average Spearman's correlation coefficients between gold and predict probing features for the two classes of sentences. Results %, reported in Table \ref{tab:sent_length} (\textbf{vediamo se tenerla}), 
showed that, for all the considered language pairs, the LinearSVR achieved higher accuracy for the probing tasks computed with respect to the incorrectly NLI classified sentences.  %($0.25$ vs. $0.23$ in average for all the linguistic tasks).
%correctly classified sentences are less correlated with sentence length ($\rho=0.23$ in average), while incorrectly classified sentences are more correlated ($\rho=0.25$ in average). 
%{\bf DOMI: secondo me qui non è molto chiaro. Ho scritto un appunto in chat!}
This is an additional evidence that deeper linguistic knowledge is needed for BERT to correctly classify the L1 of a sentences.% and that this kind of knowledge is not necessarily related to the length of the sentence (that possibly makes the sentence more complex).

%\begin{comment}
\begin{table}[h]
\small
\centering
\begin{tabular}{lllllllllll}
\hline
\textbf{Model}     & \textbf{ARA}   & \textbf{CHI}   & \textbf{TUR}   & \textbf{SPA}   & \textbf{GER}   & \textbf{FRE}   & \textbf{JPN}   & \textbf{KOR}   & \textbf{TEL}   & \textbf{HIN}   \\
\hline
Correct   & 0.226 & 0.225 & 0.236 & 0.223 & 0.215 & 0.224 & 0.276 & 0.239 & 0.234 & 0.229 \\
Incorrect & 0.248 & 0.251 & 0.249 & 0.235 & 0.244 & 0.239 & 0.290 & 0.255 & 0.258 & 0.257 \\
\hline
\end{tabular}
\caption{Average $\rho$ scores for sentences correctly and incorrectly classified using only sentence length as input feature.}
\label{tab:sent_length}
\end{table}
%\end{comment}

%Moreover, we notice that if we compute the mean sentence length of the samples correctly and incorrectly classified, the sentences 

%This is also suggested by the fact that if we compute the mean sentence length distinguishing between sentences correctly and incorrectly classified, we notice that sentences

%Interestingly, Figure \ref{fig:mse_plot} shows also that the correlation between implicit knowledge learned by BERT and its ability of solving downstream tasks is higher before the fine-tuning process. This seems to reflect what we already noticed in Section \ref{experiment_2}, i.e.\ that fine-tuning the pre-trained model on a specific task slightly degrades its competence in encoding general knowledge about the language.

\section{Conclusion}
\label{conclusion}
In this paper we studied what kind of linguistic properties are stored in the internal representations learned by BERT before and after a fine-tuning process and how this implicit knowledge correlates with the model predictions when it is trained on a specific downstream task.
Using a suite of 68 probing tasks, we showed that the pre-trained version of BERT encodes a wide range of linguistic phenomena across its 12 layers, but the order in which probing features are stored in the internal representations does not necessarily reflect the traditional division with respect to the linguistic annotation levels. We also found that BERT tends to lose its precision in encoding our set of probing features after the fine-tuning process, probably because it is storing more task--related information for solving NLI.
%QUI NON SERVE (o va spiegato meglio perché è importante): Interestingly, we noticed that features encoding verbal tense knowledge are the ones that decreases significantly for all the fine-tuned models.
%We thus think that further work needs to be done to investigate what kind of discriminant linguistic properties properties emerge after a fine-tuning process.
%This is particularly evident for the models fine-tuned on the classification of language pairs belonging to the same family (SPA-ITA, GER-ITA). 
%Se possibile Scrivere meglio:
Finally, we showed that the implicit linguistic knowledge encoded by BERT positively affects % is strongly correlated with
its ability to solve the tested downstream tasks. %In particular, we first showed that, regardless of the layer and model taken into account, most of the probing features are involved in discriminating the sentences correctly or incorrectly classified by the fine-tuned models. Second, we noticed that for such features the probing model performance show an improvement when BERT correctly predicts the L1 of a native speaker, and this is especially true for the pre--trained model. This suggests that its capacity to encode linguistic information has an influence on its predictions. % decisions. 
%In future work, we would like to extend our approach to other NLMs, such as ELMo \cite{peters2018deep} or XLNet \cite{yang2019xlnet}, and to investigate how the linguistic information implicitly encoded in such models affects different downstream tasks. 
%The demonstrated influence of linguistic competence of NLM on classification tasks would allow us to develop NLMs able to maximize 

%In future work, we plan to study how the linguistic information encoded in a NLM arise during training, performing the probing tasks on several sentence representations extracted in the pre-training phase. The aim of this investigation is studying new strategies to maximize the linguistic competence of a NLM, for example adding during the pre-training process specific linguistic tasks.

%Moreover, it would be interesting to study how the linguistic information encoded in a NLM arise and evolve as these models are trained, performing the probing tasks on several sentence representations extracted during the pre-training process. %DA FELICE: se riusciamo aggiungere anche: Un ulteriore campo di indagine sarà quello di generare NLM massimizzando la loro competenza linguistica ad esempio adding at the pre-training process specific linguistic tasks.

\bibliographystyle{acl}
\bibliography{anthology,coling2020}

\begin{thebibliography}{}

\bibitem[\protect\citename{Adi \bgroup et al.\egroup }2016]{adi2016fine}
Yossi Adi, Einat Kermany, Yonatan Belinkov, Ofer Lavi, and Yoav Goldberg.
\newblock 2016.
\newblock Fine-grained analysis of sentence embeddings using auxiliary
  prediction tasks.
\newblock {\em arXiv preprint arXiv:1608.04207}.

\bibitem[\protect\citename{Belinkov and Glass}2019]{belinkov2019analysis}
Yonatan Belinkov and James Glass.
\newblock 2019.
\newblock Analysis methods in neural language processing: A survey.
\newblock {\em Transactions of the Association for Computational Linguistics},
  7:49--72.

\bibitem[\protect\citename{Belinkov \bgroup et al.\egroup
  }2017]{belinkov2017evaluating}
Yonatan Belinkov, Llu{\'\i}s M{\`a}rquez, Hassan Sajjad, Nadir Durrani, Fahim
  Dalvi, and James Glass.
\newblock 2017.
\newblock Evaluating layers of representation in neural machine translation on
  part-of-speech and semantic tagging tasks.
\newblock In {\em Proceedings of the Eighth International Joint Conference on
  Natural Language Processing (Volume 1: Long Papers)}, pages 1--10.

\bibitem[\protect\citename{Blanchard \bgroup et al.\egroup
  }2013]{blanchard2013toefl11}
Daniel Blanchard, Joel Tetreault, Derrick Higgins, Aoife Cahill, and Martin
  Chodorow.
\newblock 2013.
\newblock Toefl11: A corpus of non-native english.
\newblock {\em ETS Research Report Series}, 2013(2):i--15.

\bibitem[\protect\citename{Blevins \bgroup et al.\egroup
  }2018]{blevins2018deep}
Terra Blevins, Omer Levy, and Luke Zettlemoyer.
\newblock 2018.
\newblock Deep rnns encode soft hierarchical syntax.
\newblock In {\em Proceedings of the 56th Annual Meeting of the Association for
  Computational Linguistics (Volume 2: Short Papers)}, pages 14--19.

\bibitem[\protect\citename{Brunato \bgroup et al.\egroup
  }2020]{profilingud-brunato-2020}
Dominique Brunato, Andrea Cimino, Felice Dell'Orletta, Giulia Venturi, and
  Simonetta Montemagni.
\newblock 2020.
\newblock Profiling-ud: a tool for linguistic profiling of texts.
\newblock In {\em Proceedings of The 12th Language Resources and Evaluation
  Conference}, pages 7147--7153, Marseille, France, May. European Language
  Resources Association.

\bibitem[\protect\citename{Cimino \bgroup et al.\egroup
  }2018]{Cimino2018SentencesAD}
Andrea Cimino, Felice Dell'Orletta, Dominique Brunato, and Giulia Venturi.
\newblock 2018.
\newblock Sentences and documents in native language identification.
\newblock In {\em CLiC-it}.

\bibitem[\protect\citename{Clark \bgroup et al.\egroup
  }2019]{clark-etal-2019-bert}
Kevin Clark, Urvashi Khandelwal, Omer Levy, and Christopher~D. Manning.
\newblock 2019.
\newblock What does {BERT} look at? an analysis of {BERT}{'}s attention.
\newblock In {\em Proceedings of the 2019 ACL Workshop BlackboxNLP: Analyzing
  and Interpreting Neural Networks for NLP}, pages 276--286, Florence, Italy,
  August. Association for Computational Linguistics.

\bibitem[\protect\citename{Conneau \bgroup et al.\egroup }2018]{conneau2018you}
Alexis Conneau, Germ{\'a}n Kruszewski, Guillaume Lample, Lo{\"\i}c Barrault,
  and Marco Baroni.
\newblock 2018.
\newblock What you can cram into a single \$\&!\#* vector: Probing sentence
  embeddings for linguistic properties.
\newblock In {\em Proceedings of the 56th Annual Meeting of the Association for
  Computational Linguistics (Volume 1: Long Papers)}, pages 2126--2136.

\bibitem[\protect\citename{Dell{'}Orletta \bgroup et al.\egroup
  }2014]{dellorletta-etal-2014-assessing}
Felice Dell{'}Orletta, Martijn Wieling, Giulia Venturi, Andrea Cimino, and
  Simonetta Montemagni.
\newblock 2014.
\newblock Assessing the readability of sentences: Which corpora and features?
\newblock In {\em Proceedings of the Ninth Workshop on Innovative Use of {NLP}
  for Building Educational Applications}, pages 163--173, Baltimore, Maryland,
  June. Association for Computational Linguistics.

\bibitem[\protect\citename{Devlin \bgroup et al.\egroup }2019]{devlin2019bert}
Jacob Devlin, Ming-Wei Chang, Kenton Lee, and Kristina Toutanova.
\newblock 2019.
\newblock Bert: Pre-training of deep bidirectional transformers for language
  understanding.
\newblock In {\em Proceedings of the 2019 Conference of the North American
  Chapter of the Association for Computational Linguistics: Human Language
  Technologies, Volume 1 (Long and Short Papers)}, pages 4171--4186.

\bibitem[\protect\citename{Goldberg}2019]{goldberg2019assessing}
Yoav Goldberg.
\newblock 2019.
\newblock Assessing bert's syntactic abilities.
\newblock {\em arXiv preprint arXiv:1901.05287}.

\bibitem[\protect\citename{Hewitt and Liang}2019]{hewitt2019designing}
John Hewitt and Percy Liang.
\newblock 2019.
\newblock Designing and interpreting probes with control tasks.
\newblock In {\em Proceedings of the 2019 Conference on Empirical Methods in
  Natural Language Processing and the 9th International Joint Conference on
  Natural Language Processing (EMNLP-IJCNLP)}, pages 2733--2743.

\bibitem[\protect\citename{Hewitt and Manning}2019]{hewitt2019structural}
John Hewitt and Christopher~D Manning.
\newblock 2019.
\newblock A structural probe for finding syntax in word representations.
\newblock In {\em Proceedings of the 2019 Conference of the North American
  Chapter of the Association for Computational Linguistics: Human Language
  Technologies, Volume 1 (Long and Short Papers)}, pages 4129--4138.

\bibitem[\protect\citename{Jain and Wallace}2019]{jain-wallace-2019-attention}
Sarthak Jain and Byron~C. Wallace.
\newblock 2019.
\newblock {A}ttention is not {E}xplanation.
\newblock In {\em Proceedings of the 2019 Conference of the North {A}merican
  Chapter of the Association for Computational Linguistics: Human Language
  Technologies, Volume 1 (Long and Short Papers)}, pages 3543--3556,
  Minneapolis, Minnesota, June. Association for Computational Linguistics.

\bibitem[\protect\citename{Jawahar \bgroup et al.\egroup
  }2019]{jawahar2019does}
Ganesh Jawahar, Beno{\^\i}t Sagot, Djam{\'e} Seddah, Samuel Unicomb, Gerardo
  I{\~n}iguez, M{\'a}rton Karsai, Yannick L{\'e}o, M{\'a}rton Karsai, Carlos
  Sarraute, {\'E}ric Fleury, et~al.
\newblock 2019.
\newblock What does bert learn about the structure of language?
\newblock In {\em 57th Annual Meeting of the Association for Computational
  Linguistics (ACL), Florence, Italy}.

\bibitem[\protect\citename{K{\'a}d{\'a}r \bgroup et al.\egroup
  }2017]{kadar2017representation}
Akos K{\'a}d{\'a}r, Grzegorz Chrupa{\l}a, and Afra Alishahi.
\newblock 2017.
\newblock Representation of linguistic form and function in recurrent neural
  networks.
\newblock {\em Computational Linguistics}, 43(4):761--780.

\bibitem[\protect\citename{Karpathy \bgroup et al.\egroup
  }2015]{karpathy2015visualizing}
Andrej Karpathy, Justin Johnson, and Li~Fei-Fei.
\newblock 2015.
\newblock Visualizing and understanding recurrent networks.
\newblock {\em arXiv preprint arXiv:1506.02078}.

\bibitem[\protect\citename{Kovaleva \bgroup et al.\egroup
  }2019]{kovaleva-etal-2019-revealing}
Olga Kovaleva, Alexey Romanov, Anna Rogers, and Anna Rumshisky.
\newblock 2019.
\newblock Revealing the dark secrets of {BERT}.
\newblock In {\em Proceedings of the 2019 Conference on Empirical Methods in
  Natural Language Processing and the 9th International Joint Conference on
  Natural Language Processing (EMNLP-IJCNLP)}, pages 4365--4374, Hong Kong,
  China, November. Association for Computational Linguistics.

\bibitem[\protect\citename{Li \bgroup et al.\egroup }2016]{li2016visualizing}
Jiwei Li, Xinlei Chen, Eduard Hovy, and Dan Jurafsky.
\newblock 2016.
\newblock Visualizing and understanding neural models in nlp.
\newblock In {\em Proceedings of the 2016 Conference of the North American
  Chapter of the Association for Computational Linguistics: Human Language
  Technologies}, pages 681--691.

\bibitem[\protect\citename{Lin \bgroup et al.\egroup }2019]{lin-etal-2019-open}
Yongjie Lin, Yi~Chern Tan, and Robert Frank.
\newblock 2019.
\newblock Open sesame: Getting inside {BERT}{'}s linguistic knowledge.
\newblock In {\em Proceedings of the 2019 ACL Workshop BlackboxNLP: Analyzing
  and Interpreting Neural Networks for NLP}, pages 241--253, Florence, Italy,
  August. Association for Computational Linguistics.

\bibitem[\protect\citename{Liu \bgroup et al.\egroup
  }2019]{liu-etal-2019-linguistic}
Nelson~F. Liu, Matt Gardner, Yonatan Belinkov, Matthew~E. Peters, and Noah~A.
  Smith.
\newblock 2019.
\newblock Linguistic knowledge and transferability of contextual
  representations.
\newblock In {\em Proceedings of the 2019 Conference of the North {A}merican
  Chapter of the Association for Computational Linguistics: Human Language
  Technologies, Volume 1 (Long and Short Papers)}, pages 1073--1094,
  Minneapolis, Minnesota, June. Association for Computational Linguistics.

\bibitem[\protect\citename{Lubetich and Sagae}2014]{lubetich2014}
Shannon Lubetich and Kenji Sagae.
\newblock 2014.
\newblock Data-driven measurement of child language development with simple
  syntactic templates.
\newblock In {\em Proceedings of COLING 2014, the 25th International Conference
  on Computational Linguistics: Technical Papers}, pages 2151--2160.

\bibitem[\protect\citename{Malmasi \bgroup et al.\egroup
  }2017]{malmasi-etal-2017-report}
Shervin Malmasi, Keelan Evanini, Aoife Cahill, Joel Tetreault, Robert Pugh,
  Christopher Hamill, Diane Napolitano, and Yao Qian.
\newblock 2017.
\newblock A report on the 2017 native language identification shared task.
\newblock In {\em Proceedings of the 12th Workshop on Innovative Use of {NLP}
  for Building Educational Applications}, pages 62--75, Copenhagen, Denmark,
  September. Association for Computational Linguistics.

\bibitem[\protect\citename{Marvin and Linzen}2018]{marvin2018targeted}
Rebecca Marvin and Tal Linzen.
\newblock 2018.
\newblock Targeted syntactic evaluation of language models.
\newblock In {\em Proceedings of the 2018 Conference on Empirical Methods in
  Natural Language Processing}, pages 1192--1202.

\bibitem[\protect\citename{Miaschi and
  Dell{'}Orletta}2020]{miaschi-dellorletta-2020-contextual}
Alessio Miaschi and Felice Dell{'}Orletta.
\newblock 2020.
\newblock Contextual and non-contextual word embeddings: an in-depth linguistic
  investigation.
\newblock In {\em Proceedings of the 5th Workshop on Representation Learning
  for NLP}, pages 110--119, Online, July. Association for Computational
  Linguistics.

\bibitem[\protect\citename{Miaschi \bgroup et al.\egroup
  }2020]{miaschi-etal-2020}
Alessio Miaschi, Sam Davidson, Dominique Brunato, Felice Dell{'}Orletta, Kenji
  Sagae, Claudia~Helena Sanchez-Gutierrez, and Giulia Venturi.
\newblock 2020.
\newblock Tracking the evolution of written language competence in l2 spanish
  learners.
\newblock In {\em Proceedings of the Fifteenth Workshop on Innovative Use of
  NLP for Building Educational Applications}. Association for Computational
  Linguistics, July.

\bibitem[\protect\citename{Nivre \bgroup et al.\egroup
  }2016]{nivre2016universal}
Joakim Nivre, Marie-Catherine De~Marneffe, Filip Ginter, Yoav Goldberg, Jan
  Hajic, Christopher~D Manning, Ryan McDonald, Slav Petrov, Sampo Pyysalo,
  Natalia Silveira, et~al.
\newblock 2016.
\newblock Universal dependencies v1: A multilingual treebank collection.
\newblock In {\em Proceedings of the Tenth International Conference on Language
  Resources and Evaluation (LREC'16)}, pages 1659--1666.

\bibitem[\protect\citename{Perone \bgroup et al.\egroup
  }2018]{perone2018evaluation}
Christian~S Perone, Roberto Silveira, and Thomas~S Paula.
\newblock 2018.
\newblock Evaluation of sentence embeddings in downstream and linguistic
  probing tasks.
\newblock {\em arXiv preprint arXiv:1806.06259}.

\bibitem[\protect\citename{Peters \bgroup et al.\egroup }2018]{peters2018deep}
Matthew Peters, Mark Neumann, Mohit Iyyer, Matt Gardner, Christopher Clark,
  Kenton Lee, and Luke Zettlemoyer.
\newblock 2018.
\newblock Deep contextualized word representations.
\newblock In {\em Proceedings of the 2018 Conference of the North American
  Chapter of the Association for Computational Linguistics: Human Language
  Technologies, Volume 1 (Long Papers)}, pages 2227--2237.

\bibitem[\protect\citename{Radford \bgroup et al.\egroup
  }2018]{radford2018improving}
Alec Radford, Karthik Narasimhan, Tim Salimans, and Ilya Sutskever.
\newblock 2018.
\newblock Improving language understanding by generative pre-training.

\bibitem[\protect\citename{Raganato and Tiedemann}2018]{raganato2018analysis}
Alessandro Raganato and J{\"o}rg Tiedemann.
\newblock 2018.
\newblock An analysis of encoder representations in transformer-based machine
  translation.
\newblock In {\em Proceedings of the 2018 EMNLP Workshop BlackboxNLP: Analyzing
  and Interpreting Neural Networks for NLP}. Association for Computational
  Linguistics.

\bibitem[\protect\citename{Sanguinetti and Bosco}2015]{sanguinetti2015parttut}
Manuela Sanguinetti and Cristina Bosco.
\newblock 2015.
\newblock Parttut: The turin university parallel treebank.
\newblock In {\em Harmonization and Development of Resources and Tools for
  Italian Natural Language Processing within the PARLI Project}, pages 51--69.
  Springer.

\bibitem[\protect\citename{Saphra and Lopez}2019]{saphra2019understanding}
Naomi Saphra and Adam Lopez.
\newblock 2019.
\newblock Understanding learning dynamics of language models with svcca.
\newblock In {\em Proceedings of the 2019 Conference of the North American
  Chapter of the Association for Computational Linguistics: Human Language
  Technologies, Volume 1 (Long and Short Papers)}, pages 3257--3267.

\bibitem[\protect\citename{Silveira \bgroup et al.\egroup
  }2014]{silveira2014gold}
Natalia Silveira, Timothy Dozat, Marie-Catherine De~Marneffe, Samuel~R Bowman,
  Miriam Connor, John Bauer, and Christopher~D Manning.
\newblock 2014.
\newblock A gold standard dependency corpus for english.
\newblock In {\em LREC}, pages 2897--2904.

\bibitem[\protect\citename{Tang \bgroup et al.\egroup
  }2018]{tang-etal-2018-analysis}
Gongbo Tang, Rico Sennrich, and Joakim Nivre.
\newblock 2018.
\newblock An analysis of attention mechanisms: The case of word sense
  disambiguation in neural machine translation.
\newblock In {\em Proceedings of the Third Conference on Machine Translation:
  Research Papers}, pages 26--35, Belgium, Brussels, October. Association for
  Computational Linguistics.

\bibitem[\protect\citename{Tenney \bgroup et al.\egroup
  }2019a]{tenney-etal-2019-bert}
Ian Tenney, Dipanjan Das, and Ellie Pavlick.
\newblock 2019a.
\newblock {BERT} rediscovers the classical {NLP} pipeline.
\newblock In {\em Proceedings of the 57th Annual Meeting of the Association for
  Computational Linguistics}, pages 4593--4601, Florence, Italy, July.
  Association for Computational Linguistics.

\bibitem[\protect\citename{Tenney \bgroup et al.\egroup }2019b]{tenney2019you}
Ian Tenney, Patrick Xia, Berlin Chen, Alex Wang, Adam Poliak, R~Thomas McCoy,
  Najoung Kim, Benjamin Van~Durme, Samuel~R Bowman, Dipanjan Das, et~al.
\newblock 2019b.
\newblock What do you learn from context? probing for sentence structure in
  contextualized word representations.
\newblock {\em arXiv preprint arXiv:1905.06316}.

\bibitem[\protect\citename{van Halteren}2004]{vanHalteren:2004}
Hans van Halteren.
\newblock 2004.
\newblock Linguistic profiling for author recognition and verification.
\newblock In {\em Proceedings of the Association for Computational
  Linguistics}, pages 200--207.

\bibitem[\protect\citename{Vig and Belinkov}2019]{vig-belinkov-2019-analyzing}
Jesse Vig and Yonatan Belinkov.
\newblock 2019.
\newblock Analyzing the structure of attention in a transformer language model.
\newblock In {\em Proceedings of the 2019 ACL Workshop BlackboxNLP: Analyzing
  and Interpreting Neural Networks for NLP}, pages 63--76, Florence, Italy,
  August. Association for Computational Linguistics.

\bibitem[\protect\citename{Warstadt \bgroup et al.\egroup
  }2019]{warstadt2019investigating}
Alex Warstadt, Yu~Cao, Ioana Grosu, Wei Peng, Hagen Blix, Yining Nie, Anna
  Alsop, Shikha Bordia, Haokun Liu, Alicia Parrish, et~al.
\newblock 2019.
\newblock Investigating bert’s knowledge of language: Five analysis methods
  with npis.
\newblock In {\em Proceedings of the 2019 Conference on Empirical Methods in
  Natural Language Processing and the 9th International Joint Conference on
  Natural Language Processing (EMNLP-IJCNLP)}, pages 2870--2880.

\bibitem[\protect\citename{Zeldes}2017]{Zeldes2017}
Amir Zeldes.
\newblock 2017.
\newblock The {GUM} corpus: Creating multilayer resources in the classroom.
\newblock {\em Language Resources and Evaluation}, 51(3):581--612.

\bibitem[\protect\citename{Zhang and Bowman}2018]{zhang2018language}
Kelly Zhang and Samuel Bowman.
\newblock 2018.
\newblock Language modeling teaches you more than translation does: Lessons
  learned through auxiliary syntactic task analysis.
\newblock In {\em Proceedings of the 2018 EMNLP Workshop BlackboxNLP: Analyzing
  and Interpreting Neural Networks for NLP}, pages 359--361.

\end{thebibliography}

\end{document}